%% file: main_paper.tex
\pdfoutput=1
\documentclass{article}

\PassOptionsToPackage{numbers, compress}{natbib}

\usepackage{graphicx}
\usepackage{caption}
\usepackage{subcaption}
\usepackage{comment}
\usepackage{amsmath,amssymb} 
\usepackage[algo2e]{algorithm2e}
\usepackage{algorithm}
\usepackage{color}

\usepackage[preprint]{neurips_2022}

\usepackage[utf8]{inputenc} 
\usepackage[T1]{fontenc}    
\usepackage{hyperref}       
\usepackage{url}            
\usepackage{booktabs}       
\usepackage{amsfonts}       
\usepackage{nicefrac}       
\usepackage{microtype}      
\usepackage{xcolor}         
\usepackage{enumitem}
\usepackage{multirow}
\usepackage{authblk}
\usepackage[toc,page,header]{appendix}
\usepackage{minitoc}
\addcontentsline{toc}{section}{Appendix} 

\newtheorem{definition}{Definition}[section]
\newtheorem{theorem}{Theorem}[section]
\newtheorem{lem}{Lemma}[section]

\newcommand{\mat}[1]{\mathbf{#1}}
\newcommand{\vect}[1]{\mathbf{#1}}
\newcommand{\vectorize}[1]{\text{vec}\left(#1\right)}
\newcommand{\norm}[1]{\left\|#1\right\|}

\newcommand{\calR}{\mathcal{R}}

\def\R{\mathbb{R}}

\def\cD{\mathcal{D}}

\def\cF{\mathcal{F}}

\def\cL{\mathcal{L}}

\def\cN{\mathcal{N}}
\def\cP{\mathcal{P}}

\newcommand{\expect}{\mathbb{E}}
\newcommand{\poly}{\mathrm{poly}}

\newcommand{\sw}[1]{{\color{orange}{\bf\sf [SW: #1]}}}

\title{Toward Sustainable Continual Learning: Detection and Knowledge Repurposing of Similar Tasks}

\author[1]{Sijia Wang}
\author[2]{Yoojin Choi}
\author[1]{Junya Chen}
\author[2]{Mostafa El-Khamy}
\author[1]{Ricardo Henao}

\affil[1]{Duke University}
\affil[2]{SoC R\&D Samsung Semiconductor Inc.}
\affil[ ]{\texttt {sijia.wang@duke.edu}}

\begin{document}

\maketitle

\begin{abstract}
Most existing works on continual learning (CL) focus on overcoming the catastrophic forgetting (CF) problem, with dynamic models and replay methods performing exceptionally well. However, since current works tend to assume exclusivity or dissimilarity among learning tasks, these methods require constantly accumulating task-specific knowledge in memory for each task. This results in the eventual prohibitive expansion of the knowledge repository if we consider learning from a long sequence of tasks.
In this work, we introduce a paradigm where the continual learner gets a sequence of mixed similar and dissimilar tasks.
We propose a new continual learning framework that uses a task similarity detection function that does not require additional learning, with which we analyze whether there is a specific task in the past that is similar to the current task.
We can then reuse previous task knowledge to slow down parameter expansion, ensuring that the CL system expands the knowledge repository sublinearly to the number of learned tasks.
Our experiments show that the proposed framework performs competitively on widely used computer vision benchmarks such as \texttt{CIFAR10}, \texttt{CIFAR100}, and \texttt{EMNIST}.
\end{abstract}

\section{Introduction}\label{introduction}
Human intelligence is distinguished by the ability to learn new tasks over time while remembering how to perform previously experienced tasks. Continual learning (CL), an advanced machine learning paradigm requiring intelligent agents to continuously learn new knowledge while trying not to forget past knowledge, has a pivotal role in machines imitating human-level intelligence \cite{Hassabis2017NeuroIns}. The main problem in continual learning is catastrophic forgetting (CF) of previous knowledge when new tasks, observed over time, are incorporated to the model via training or (parameter) expansion.

In general, continual learning methods can be categorized into three major groups based on: ($i$) regularization, ($ii$) replay, and ($iii$) dynamic (expansion) models. Approaches based on regularization \cite{Kirkpatrick2017OvercomingCF,ucl2019, Nguyen2018VariationalCL} alleviate forgetting by constraining the updates of parameters that are important for previous tasks, by adding a regularization term to the loss function penalizing changes between parameters for current and previous tasks. While this type of approach excels at keeping a low memory footprint, the ability to remember previous knowledge eventually declines, especially, in scenarios with long sequences of tasks \cite{Schwarz2018ProgressC}. Methods based on replay keep a memory bank to store either a subset of exemplars to represent each class and task \cite{Rebuffi2017iCaRLIC,Rolnick2019ExperienceRF,Isele2018SelectiveER}, or generative models for pseudo-replay \cite{shin2017,vandeVen2018GenerativeRW}. Approaches based on dynamic models  \cite{Li2019LearnTG, Lee2020AND,yoon2018,Rusu2016ProgressiveNN,Vniat2021EfficientCL,Ostapenko2021ContinualLV} allow the architecture (or components of it) to expand over time. Specifically, \cite{Vniat2021EfficientCL,Ostapenko2021ContinualLV} leverage modular networks and study a practical continual learning scenario with 100-task long sequences. Nevertheless, these parameter expansion approaches face the challenges of keeping model growth under control (ideally sublinear). This problem associated with parameter expansion in CL systems is critical and requires further attention.

This paper discusses CL under a task continual learning (TCL) setting, \emph{i.e.}, that in which data arrives sequentially in groups of \emph{tasks}.
For works under this scenario \cite{Hsu2019,Delange_2021}, the assumption is usually that once a new task is presented, all of its data becomes readily available for batch (offline) training. In this setting, a \emph{task} is defined as an individual training phase with a new collection of data that belongs to a new (never seen) group of classes, or in general, a new domain. Further, TCL also (implicitly) requires a task identifier during training.
However, in practice, when the model has seen enough tasks, a newly arriving batch of data becomes increasingly likely to belong to the same group of classes or domain of a previously seen task. Importantly, most existing works on TCL fail to acknowledge this possibility. Moreover and in general, the task definition or identifier may not be available during training, {\em e.g.}, the model does not have access to the task description due to (user) privacy concerns.
In such case mostly concerning dynamic models, the system has to treat every task as new, thus constantly learning new sets of parameters regardless of task similarity or overlap.
This clearly constitutes a suboptimal use of resources (predominantly memory), especially as the number of tasks experienced by the CL system grows.

This study investigates the aforementioned scenario and makes it an endeavor to create a memory-efficient CL system which though focused on image classification tasks, is general and in principle can be readily used toward other applications or data modality settings. We provide a solution for dynamic models to identify similar tasks when no task identifier is provided during the training phase. To the best of our knowledge, the only work that also discusses the learning of a continual learning system with mixed similar and dissimilar tasks is \cite{mixedsequence}, which proposes a task similarity function to identify previously seen similar tasks, which requires training a reference model every time a new task becomes available.
Alternatively, in this work, we identify similar tasks \emph{without} the need for training a new model, by leveraging a task similarity metric, which in practice results in high task similarity identification accuracy. We also discuss memory usage under challenging scenarios where longer, more realistic, sequences of more than 20 tasks are used.

To summarize, our contributions are listed below:
\begin{itemize}[leftmargin=4mm]
    \item We propose a new framework for an under-explored and yet practical TCL setting in which we seek to learn a sequence of mixed similar and dissimilar tasks, while preventing (catastrophic) forgetting and repurposing task-specific parameters from a previously seen similar task, thus slowing down parameter expansion.
    \item The proposed TCL framework is characterized by a task similarity detection module that determines, without additional learning, whether the CL system can reuse the task-specific parameters of the model for a previous task or needs to instantiate new ones. 
    \item Our task similarity detection module shows remarkable performance on widely used computer vision benchmarks, such as \texttt{CIFAR10} \cite{krizhevsky2009learning}, \texttt{CIFAR100} \cite{krizhevsky2009learning}, \texttt{EMNIST} \cite{Cohen2017EMNISTAE}, from which we create sequences of 10 to 100 tasks.
\end{itemize}

\section{Related Work}
\label{Related Work}
\label{CL literature review}
\textbf{TCL in Practical Scenarios} Task continual learning (TCL) being an intuitive imitation of human learning process constitutes one of the most studied scenarios in CL. Though TCL systems have achieved impressive performance \cite{Serr2018OvercomingCF, Kaushik2021UnderstandingCF}, previous works have mainly focused on circumventing the problems associated with CF. Historically, task sequences have been restricted to no more than 10 tasks and strong assumptions have been imposed so that all the tasks in the sequence are unique and classes among tasks are disjoint \cite{vandeVen2018GenerativeRW, vandeVen2019ThreeSF}. The authors of \cite{prabhu2020greedy} rightly argue that currently discussed CL settings are oversimplified, and more general and practical CL forms should be discussed to advance the field. It is not until recently that solutions have been proposed for longer sequences and more practical CL scenarios. Particularly, \cite{mixedsequence} proposed CAT, which learns from a sequence of mixed similar and dissimilar tasks, thus enabling knowledge transfer between future and past tasks detected to be similar. 
To characterize the tasks, a set of task-specific masks, {\em i.e.}, binary matrices indicating which parameters are important for a given task \cite{Serr2018OvercomingCF}, are trained along other model parameters. 
Specifically, these masks are activated and parameters associated to them are finetuned once the current task is identified as ``similar'' by a task similarity function, or otherwise held fixed by the masking parameters to protect them from changing, hence preventing CF.
Alternatively, \cite{Vniat2021EfficientCL} introduces a modular network, which is composed of neural modules that are potentially shared with related tasks. Each task is optimized by selecting a set of modules that are either freshly trained on the new task or borrowed from similar past tasks. These similar tasks are detected by a task-driven prior.
\cite{Vniat2021EfficientCL, Ostapenko2021ContinualLV} evaluate their approach with a real-world CL benchmark CtrL that contains 100 tasks with class overlap between tasks.
The data-driven prior used in \cite{Vniat2021EfficientCL} for recognizing similar tasks is a very simple approach, which leaves space for improved substitutes yielding slower parameter growth. In summary, these works serve as a reminder that identifying similar tasks in TCL settings is in general a hard problem that deserves more attention.

\textbf{Computational Efficiency in CL} The first and foremost step for tackling computational inefficiencies is to identify the framework components that are most computationally consuming. For instance, to avoid storing real images for replay, \cite{Iscen2020MemoryEfficientIL} proposes to keep lower-dimensional feature embeddings of images instead. Another way to tackle the issue is through limiting trainable parameters in the model architecture itself by partitioning a convolutional layer into a backbone and task-specific adapting modules, like in \cite{varshney2021camgan, cong2020gan}. Similarly, Filter Atom Swapping in \cite{miao2022continual} decompose filters in neural network layers and ensemble them with filters of tasks in the past. We can also make use of task relevancy or similarity. For instance, approaches with model structures similar to Progressive Network \cite{Rusu2016ProgressiveNN}, where tasks are optimized in search of neural modules to use. Further, \cite{Vniat2021EfficientCL} is proposed to optimize search space through task similarity.

\section{Background}\label{background}
\subsection{Problem Setting}\label{problem setting}
We consider the TCL scenario for image classification tasks, where we seek to incrementally learn a sequence of tasks {$\mathcal{T}_1, \mathcal{T}_2, \ldots, \mathcal{T}_{t-1}$} and denote the collection of tasks currently learned as ${\cal T}=\{{\cal T}_1,\ldots,{\cal T}_{t-1}\}$. The underlying assumption is that, as the number of tasks in $\mathcal{T}$ grows, the current task $\mathcal{T}_t$ will eventually have a corresponding similar task $\mathcal{T}^{sim}_t \in \mathcal{T}$. Let the set of all dissimilar tasks be $\mathcal{T}^{dis}_t = {\cal T}\backslash \mathcal{T}_t^{sim}$. We define similar and dissimilar tasks as follows.

\textit{Similar} and \textit{Dissimilar} tasks: Consider two tasks $A$ and $B$, which are represented by datasets $D_{A} = \{\vect{x}_{i}^{A}, y^{A}_{i}\}^{n_A}_{i=1}$ and $D_{B} = \{\mathbf{x}_{i}^{B}, y^{B}_{i}\}^{n_B}_{i=1}$, where $y_{i}^{A} \in \{\mathcal{Y}^A\}$ and $y_{i}^{B} \in \{ \mat{Y}^B\}$. If predictors (\emph{e.g.}, images) in $\{\mathbf{x}_{i}^{A}\}^{n_A}_{i=1} \sim \cP$ and $\{\mathbf{x}_{i}^{B}\}^{n_B}_{i=1} \sim \mathcal{P}$, and labels $\{\mathcal{Y}^A\} = \{\mathcal{Y}^B\}$, indicating that data from $A$ and $B$ belong to the same group of classes, and their predictors are drawn from the same distribution $\mathcal{P}$, then we say A and B are \emph{similar} tasks, otherwise, $A$ and $B$ are deemed \emph{dissimilar}.
Notably, when both tasks share the same distribution $\mathcal{P}$, but have different label spaces, they are considered dissimilar.

The main objective in this work is to identify $\mathcal{T}^{sim}_{t}$ among $\mathcal{T}$ without training a new model (or learning parameters) for $\mathcal{T}_{t}$, by leveraging a task similarity identification function, that will enable the system to reuse the parameters of $\mathcal{T}^{sim}_{t}$ when identification of a previously seen task is successful.
Alternatively, the system will instantiate new parameters for the dissimilar task.
As a result, the system will attempt to learn parameters for the set of unique tasks, which in a long sequence of tasks is assumed to be smaller than the sequence length.
In practice, in order to handle memory efficiently, we do not have to instantiate completely different sets of parameters for every unique task, but rather we define \emph{global} and \emph{task-specific} parameters as a means to further control model growth.
For this purpose, we leverage the efficient feature transformations for convolutional models described below.

\subsection{Task-Specific Adaptation via Feature Transformation Techniques}
\label{feature transformation}
The efficient feature transformation (EFT) framework in \cite{Verma2021EfficientFT} proposed that instead of finetuning all the parameters in a well-trained (pretrained) model, one can instead partition the network into a (global) \emph{backbone} model and \emph{task-specific} feature transformations. 
Note that similar ideas have also been explored in \cite{cong2020gan, miao2022continual, Howard2017MobileNetsEC, Perez2018FiLMVR, Hu2018SqueezeandExcitationN}. 
Given a trained \emph{backbone} convolutional neural network, we can transform the convolutional feature maps $F$ for each layer into task-specific feature maps $W$ by implementing small convolutional transformations.

In our setting, only $W$ is learned for task $\mathcal{T}_{t}$ as a means to reduce the parameter count, and thus the memory footprint, required for new tasks.
Specifically, the feature transformation involves two types of convolutional kernels, namely, $\omega^s \in \R^{3 \times 3 \times a}$ for capturing spatial features within groups of channels and $\omega^d \in \R^{1 \times 1 \times b}$ for capturing features across channels at every location in $F$, where $a$ and $b$ are hyperparameters controlling the size of each feature map groups. 

The transformed feature maps $W = W^s + \gamma W^d$ are obtained from $W^s = [W^s_{0: a-1} | \cdots | W^s_{(K-a):K}]$ and $W^d = [W^s_{0: b-1} | \cdots | W^s_{(K-b):K}]$ via
\begin{align}
    W^{s}_{ai:(ai+a-1)} &= [\omega^s_{i,1} * F_{ai:(ai+a-1)} | \cdots | w^{s}_{i,a} * F_{ai:(ai+a-1)}], & i & \in \{0, \cdots, \frac{K}{a}-1\} \\
    W^{d  }_{bi:(bi+b-1)} &= [\omega^d_{i,1} * F_{bi:(bi+b-1)} | \cdots | w^{d}_{i,b} * F_{bi:(bi+b-1)}], & i & \in \{0, \cdots, \frac{K}{b}-1\} 
\end{align}
where the feature maps $F \in \R^{M\times N \times K}$ and $W \in \R^{M\times N \times K}$ have spatial dimensions $M$ and $N$, $K$ is the number of feature maps, | is the concatenation operation, $K/a$ and $K/b$ are the number of groups into which $F$ is split for each feature map, $\gamma \in \{0, 1\}$ indicates whether the point-wise convolutions $\omega^d$ are employed, and $W^{s}_{ai:(ai+a-1)}  \in \R^{M \times N \times a}$, and $W^{d}_{bi:(bi+b-1)}  \in \R^{M \times N \times b}$ are slices of the transformed feature map.
In practice, we set $a\ll K$ and $b\ll K$ so that the amount of trainable parameters per task is substantially reduced. For instance, using a ResNet18 backbone, $a=8$, and $b=16$ results in 449k parameters per new tasks, which is $3.9\%$ the size of the backbone.
As empirically demonstrated in \cite{Verma2021EfficientFT}, EFT preserves the remarkable representation learning power of ResNet models while significantly reducing the number of trainable parameters per task.

\vspace{-5pt}
\subsection{Task Continual Learning: Mixture Model Perspective}
\label{DPMM}
\vspace{-3pt}
One of the key components when trying to identify similar tasks is to decide whether $\{\mathbf{x}_{i}^{A}\}^{n_A}_{i=1}$ and $\{\mathbf{x}_{i}^{B}\}^{n_B}_{i=1}$ originate from the same distribution $\mathcal{P}$.
However, though conceptually simple, it is extremely challenging in practice, particularly when predictors are complex instances such as images.
Intuitively, for a sequence of tasks ${\cal T}=\{{\cal T}_1,\ldots,{\cal T}_{t-1}\}$, with corresponding data $D_1, \ldots, D_{t-1}$, $D_t = \{\mathbf{x}_i , y_i \}^{n_t}_{i=1}$ consisting of $n_t$ instances, where $\mathbf{x}_i$ is an image and $y_i$ is its corresponding label, we can think of data instances $\mathbf{x}$ from the collection of all unique tasks as a mixture model defined as
\vspace{-5pt}
\begin{align}\label{eq:dp}
    p(\mathbf{x}) = \pi_* p(\mathbf{x}|\phi_*) + \sum_{j=1}^{t-1} \pi_j p(\mathbf{x}|\phi_j) ,
\end{align}
from which we can see that $\pi_j$ is the probability that $\mathbf{x}$ belongs to task ${\cal T}_j$ and $p(\mathbf{x}|\phi_j)$ is the likelihood of $\mathbf{x}$ under the distribution for task ${\cal T}_j$ parameterized by $\phi_j$.
Further, $\pi_*$ and $\phi_*$ denote the hypothetical probability and parameters for a new \emph{unseen} task $*$, \emph{i.e.}, distinct from $\{p(\mathbf{x}|\phi_j)\}_{j=1}^{t-1}$.
The formulation in \eqref{eq:dp} which is reminiscent of a Dirichlet Process Mixture Model (DPMM) \cite{Gershman2011ATO,VIDPM}, can in principle be used to estimate \emph{a posteriori} that $p(D_t\in {\cal T}_*)$ by evaluating \eqref{eq:dp}, which assumes that parameters $\{\pi_*,\pi_1,\ldots,\pi_{t-1}\}$ and $\{\phi_*,\phi_1\ldots,\phi_{t-1}\}$ are readily available.
Unfortunately, though for the collection of existing tasks ${\cal T}$ we can effectively estimate $\pi_j$ and $p(\mathbf{x}|\phi_j)$, using generative models (\emph{e.g.}, variational autoencoers), the parameters for a new task $\phi_*$ and $p(\mathbf{x}|\phi_*)$ are much more difficult to estimate because $i$) if we naively build a generative model for the new dataset $D_t$ to obtain $\phi_t$ and then evaluate \eqref{eq:dp}, it is almost guaranteed that $D_t$ corresponding to ${\cal T}_t$ will be more likely under $p(\mathbf{x}|\phi_*=\phi_t)$ than any other existing task distribution $\{p(\mathbf{x}|\phi_j)\}_{j=1}^{t-1}$, and alternatively, $ii$) if we set $p(\mathbf{x}|\phi_*)$ to some prior distribution, \emph{e.g.} a pretrained generative model, it will be most definitely never selected, especially in scenarios with complex predictors such as images.
In fact, in early stages of development we empirically verified this being the case using both pretrained generative models admitting (marginal) likelihood estimation \cite{Higgins2017betaVAELB, Burgess2018UnderstandingDI}
and anomaly detection models based on density estimators \cite{SVDDrevisited, pmlr-v80-ruff18a}.

In Section~\ref{Method}, we will show how to leverage \eqref{eq:dp} to identify new tasks in the context of TCL without using data from ${\cal T}_t$ for learning nor specifying a prior distribution for $p(\mathbf{x}|\phi_*)$.

\vspace{-5pt}
\subsection{Estimating the Association between Predictors and Labels} \label{complexity measure}
\vspace{-3pt}
The mixture model perspective for TCL in \eqref{eq:dp} offers a compelling way to compare the distribution of predictors for different tasks, however, it does not provide any insights about the strength of the association between predictors and labels for dataset $D_t$ corresponding to task ${\cal T}_t$.
Further, \cite{zhang2021understanding} showed that overparameterized neural network classifiers can attain zero training error, regardless of the strength of association between predictors and labels, and in an extreme case, even for randomly labeled data.
However, it is clear that a model trained with random labels will not generalize.

So motivated, \cite{arora2019finegrained} first studied the properties of suitably labeled data that control generalization ability, and proposed a generalization bound on the test error (empirical risk) for arbitrary overparameterized two-layer neural network classifiers with rectified linear units (ReLU).
Importantly, unlike previous works on generalization bounds for neural networks \cite{Dziugaite2017ComputingNG, Zhou2019NonvacuousGB, Li2018OnTG}, their bound can be effectively calculated \emph{without} training the network or making assumptions about its size (number of hidden units).
Their \emph{complexity measure}, which is shown in \eqref{equation: complexity measure}, is set to directly quantify the strength of the association between data and labels without learning.
More precisely, for dataset $D=\{\mathbf{x}_i,y_i\}_{i=1}^n$ of size $n$, the generalization bound in \eqref{equation: complexity measure} is an upper bound on the (test) error conditioned on $D$,
\begin{align}\label{equation: complexity measure}
    S(D) = \sqrt{\frac{2\mathbf{y}^\mathsf{T}\mathbf{H}^{-1}\mathbf{y}}{n}}, 
\end{align}
where $\mathbf{y} = (y_1, \cdots, y_n)^\mathsf{T}$ and matrix $\mathbf{H} \in \R^{n \times n}$, which can be seen as a Gram matrix for a ReLU activation function is defined as
\begin{equation}\label{eq: Gram Matrix}
\begin{aligned}
    H_{ik} &= \mathbb{E}_{\mathbf{w} \sim \mathcal{N}(\mathbf{0}, \mathbf{I})} \big[ \mathbf{x}_{i}^{\top} \mathbf{x}_k \mathbb{I} \{ \mathbf{w}^{\top} \mathbf{x}_i \geq 0, \mathbf{w}^{\top} \mathbf{x}_k \geq 0\}\big] \\
    &= \frac{\mathbf{x}^{\top}_{i} \mathbf{x}_{k} (\pi - \arccos{(\mathbf{x}^{\top}_{i} \mathbf{x}_{k})})}{2\pi}, \ \ \forall i, k \in 1,\ldots,n ,
\end{aligned}
\end{equation} 
where $H_{ik}$ is the $(i, k)$-th entry of $\mathbf{H}$, $\mathcal{N}(0, \mathbf{I})$ denotes the standard Gaussian distribution, $\mathbf{w}$ is a weight vector in the first layer of the two-layer neural network, and $\mathbb{I}\{\cdot\}$ the an indicator function.
Empirically, \cite{arora2019finegrained} showed that the complexity measure in \eqref{equation: complexity measure} can distinguish between strong and weak associations between predictors and labels.
Effectively, weak associations tend to be consistent with \emph{randomly labeled} data, thus unlikely to generalize.

In Section~\ref{Method} we will leverage \eqref{equation: complexity measure} as a metric for similar task detection, which we will recast a measure to quantify the association between the labels for a given task and the features from encoders learned from previously seen tasks. 

\begin{figure}[t]
\centering
\includegraphics[width=\textwidth]{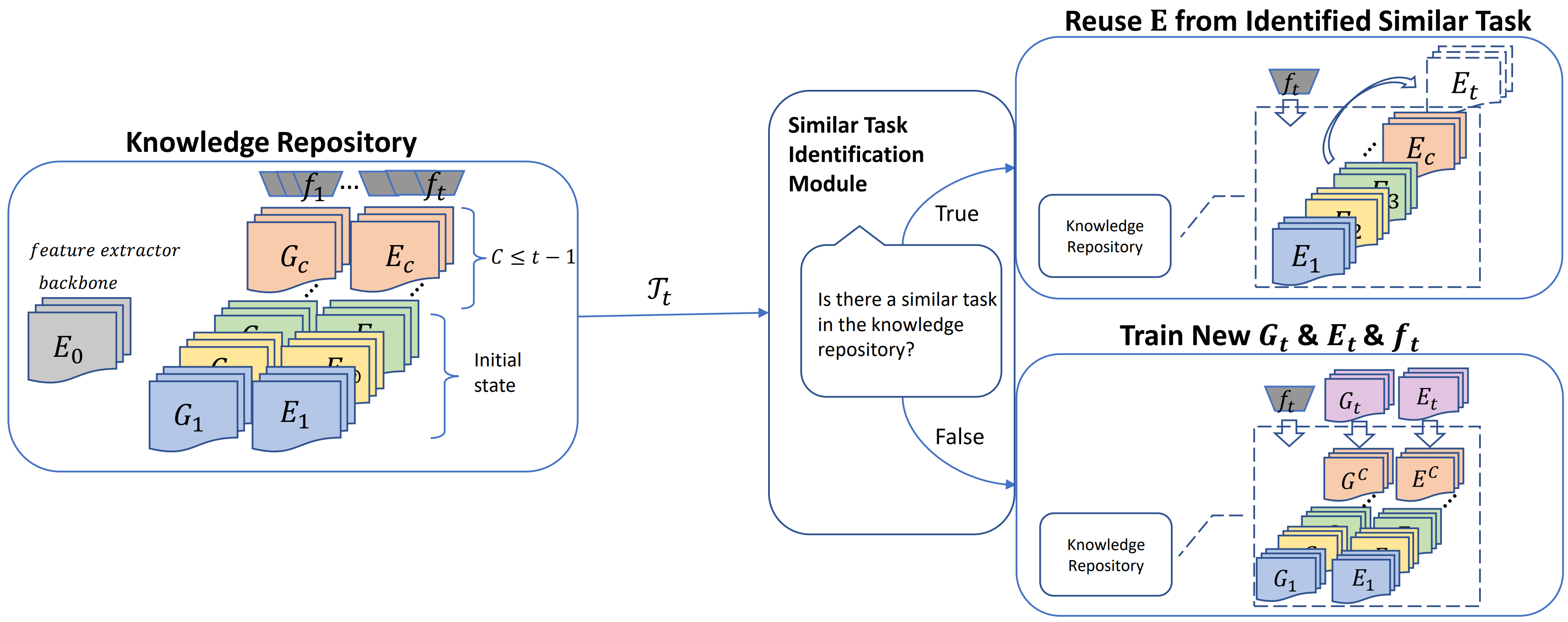}
\caption{SDR Framework: We start with a pretrained feature extractor backbone $E_0(\cdot)$ and models for three dissimilar tasks kept in a knowledge repository, so the following tasks have some tasks to compare to at first. Autoencoding pairs $G(\cdot)$ and $E(\cdot)$ are used for building the distributional consistency estimator and analyzing predictor-label association in the similar task identification module. The similar task identification module uses \eqref{eq: j(u_n = k|x_n)} and \eqref{eq: multiclass} in the middle block with training dataset $D_t$ for each new task $\mathcal{T}_t$ in the CL sequences. If there is a similar task in the knowledge repository, then simply map the new task to the identified previous task, reuse the feature extractor, $E(\cdot)$, and only learn a new classification head $f_t(\cdot)$ for the current task; alternatively, we learn a new encoder, decoder and classification head. At the end of the learning for each task, the knowledge repository is updated with the newly trained parameters. $C$ denotes the number of unique tasks in the repository, which is smaller than the total number of tasks seen so far.}

\label{fig:SDR Diagram}
\end{figure}
\vspace{-5pt}
\section{Detection and Repurposing of Similar Tasks}
\label{Method}
\vspace{-3pt}
We now propose the \textbf{S}imilar task \textbf{D}etection and \textbf{R}epurposing (\textbf{SDR}) framework for task continual learning, which though tailored to image classification tasks, can be in principle reused for other modalities, \emph{e.g,}, text and structured data.
Specifically, the framework consists of the following two components: ($i$) task-specific encoders $\{E_j(\cdot)\}^{t-1}_{j=1}$ for $\mathcal{T}=\{{\cal T}_j\}_{j=1}^{t-1}$; and ($ii$) a mechanism for similar task identification structured as two separate but complementary components, namely, a measure of distributional similarity between the current task ${\cal T}_t$ and each of all previous  tasks ${\cal T}$, and a predictor-to-label association similarity, each of which leverage the mixture model perspective and the complexity measure introduced in Sections \ref{DPMM} and \ref{complexity measure}, respectively.
For the former we use task-specific generative models $\{G_j(\cdot)\}^{t-1}_{j=1}$ specified as variational autoencoders that admit (marginal) likelihood estimation via the evidence lower bound (ELBO) \cite{Kingma2014AutoEncodingVB, Blei2016VariationalIA}, and for the latter we use the measure $S(\cdot)$ in \eqref{equation: complexity measure}.
The complete framework described below is briefly illustrated in \ref{fig:SDR Diagram} and presented as Algorithm \ref{alg:SDR} in the Supplementary Material (SM).
In a nutshell, similar task identification results in one of two outcomes, namely, a previously seen task is identified as similar to ${\cal T}_t$, in which case, we will use their corresponding encoder \emph{as is}, but will learn a task-specific classification head. Alternatively, if ${\cal T}_t$ is deemed as a new unseen task, we will learn both a task-specific encoder, decoder (generator), and classification head. The classification head for ${\cal T}_t$ is defined as $p(y|\mathbf{x})=f_t(E_t(\mathbf{x}))$ and specified as a fully connected network.

\vspace{-5pt}
\subsection{Task-specific Encoder}
\label{task-specific-encoder}
\vspace{-3pt}
Following the specifications for the efficient feature transformations in \cite{Verma2021EfficientFT}, we first specify a pretrained backbone encoder $E_{0}(\cdot)$, then for each new unseen task, we adapt $E_{0}(\cdot)$ into ${E}_{t}(\cdot)$ using an EFT module that is learned together (end-to-end) with the task-specific classification head $p(y|\mathbf{x})=f_t(E_t(\mathbf{x}))$ in a supervised fashion using $D_t$, while keeping $E_{0}(\cdot)$ fixed. 
\vspace{-5pt}
\subsection{Similar Task Identification}
\label{sec: search-engine}
\vspace{-3pt}
As previously briefly described, the procedure to identify similar tasks for a new task ${\cal T}_t$ among existing ${\cal T}=\{{\cal T}_j\}_{j=1}^{t-1}$, amounts to estimate the distributional similarity of the predictors and the strength of the association between labels and predictors.

Provided a sequence of tasks ${\cal T}$ with corresponding encoders and generators $\{E_j(\cdot)\}^{t-1}_{j=1}$ and $\{G_j(\cdot)\}^{t-1}_{j=1}$, respectively, we can use \eqref{eq:dp} to estimate the likelihood that predictors in $D_t$ are consistent with the generator for a previously seen task.
Specifically,
\begin{align}\label{eq: j(u_n = k|x_n)}
    p({\cal P}_t={\cal P}_j) = \mathbb{E}_{\mathbf{x}_i\in D_t}[p(u_i = j|\mathbf{x}_i)] = \frac{1}{n}\sum_{i=1}^{n} \frac{\pi_{j} p(\mathbf{x}_i| \phi_j)}{\sum_{j^{'}} \pi_{j^{'}} p(\mathbf{x}_i| \phi_{j^{'}})} ,
\end{align}
where in a slight abuse of notation we let $p({\cal P}_t={\cal P}_j)$ indicate the probability that ${\cal P}_t$ and ${\cal P}_j$ have the same predictor distributions.
In practice, we use the expected lower bound (ELBO) of a variational autoencoder pair $E_m(\cdot)$ and $G_m(\cdot)$ to approximate the (marginal) likelihood $p(\mathbf{x}_i| \phi_j)$ parameterized by $\phi_j$, which encapsulates the parameters of both $E_j(\cdot)$ and $G_j(\cdot)$, \emph{i.e.}, encoder and generator, respectively.
However, $p({\cal P}_t={\cal P}_j)$ though useful for previously seen tasks ${\cal T}$, does not help to identify new unseen tasks that will be consistent with a hypothetical distribution ${\cal P}_*$ specified as $p(\mathbf{x}_i| \phi_*)$ and parameterized by $\phi_*$.
Intuitively, given a new unseen task ${\cal T}_t$, $p({\cal P}_t={\cal P}_j)$ is likely to be a drawn from a uniform distribution in $t-1$ dimensions, so when $p({\cal P}_t={\cal P}_j)\approx 1/(t-1)$, for $j=1,\ldots,t-1$, the predictors for the new task are likely from a new unseen task.
Alternatively, when  $p({\cal P}_t={\cal P}_j)\to 1$, for some task ${\cal T}_j$, the predictors in $D_t$ are likely to be consistent in distribution to ${\cal P}_j$, \emph{i.e.}, to predictors from dataset $D_j$.

Importantly, the distributional consistency estimator in \eqref{eq: j(u_n = k|x_n)} only estimates the probability that predictors in $D_t$ are consistent in distribution to that of $D_j$. 
However, we still need to estimate the likelihood that encoder $E_j(\cdot)$ built from $D_j$ is strongly associated with the labels of interest in $D_t$ without building a classification model for $D_t$ using $E_j(\cdot)$.
For this purpose, we leverage the complexity measure in \eqref{equation: complexity measure}, which we write below in terms of the task-specific encoders.

Following the construction of $\mathbf{H}$ as defined in \eqref{eq: Gram Matrix}, we start by redefining it in terms of the task-specific encoders.
Specifically, given encoders $\{E_{j}(\cdot)\}^{t-1}_{j=1}$ and dataset $D_{t} = \{\mathbf{x}_{i}^{t}, y^{t}_{i}\}^{n_t}_{i=1}$ for task ${\cal T}_t$, we extract features for the predictors in $D_{t}$ using all available encoders, \emph{i.e.}, $\{\{E_{j}(\mathbf{x}_{i}^{t})\}^{n_t}_{i=1}\}^{t-1}_{j=1}$. 
For convenience, let $\mathbf{e}^{j, t}_{i} = E_{j}(\mathbf{x}^{t}_i)$.
We can rewrite the Gram matrix $\mathbf{H}^{j, t}$ corresponding to encoder $j$ and dataset $D_t$ in \eqref{eq: Gram Matrix} as
\begin{equation}\label{eq: Our Gram Matrix}
\begin{aligned}
    \mathbf{H}^{j, t}_{ik} &= \frac{(\mathbf{e}^{j, t}_{i})^\top \mathbf{e}^{j, t}_{k}(\pi - \arccos((\mathbf{e}^{j, t}_i)^\top \mathbf{e}^{j, t}_{k}))}{2\pi} ,
\end{aligned}    
\end{equation}
which is a multi-class extension of the complexity measure for binary classification tasks introduced in \cite{arora2019finegrained}. 
Note that the Gram matrix $\mathbf{H}^{j, t} \in \R^{n_t \times n_t}$ has the same formulation as for when $\mathbf{y}^t = (y^{t}_1, \cdots, y^{t}_n) \in \{0,1\}^{c \times n}$, where $c$ is the number of classes.
Let $\mathbf{A}^{j, t} = \mathbf{y}^\mathsf{T} (\mathbf{H}^{j, t})^{-1}\mathbf{y} \in \R^{c \times c}$, the similarity metric $S$ between tasks $\mathcal{T}_t$ and $\mathcal{T}_j$ via encoder $E_j(\cdot)$ is defined as:
\begin{equation} \label{eq: multiclass}
    S(j, t) = \sqrt{\frac{2 ||{\mathbf{A}^{j, t}}^\top \mathbf{A}^{j, t}||^2_2} {n_t}} ,
\end{equation}
where $||\cdot||^2_2$ is the Frobenius Norm, $n_t$ is the size of $D_t$.
The proof for \eqref{eq: Our Gram Matrix} and \eqref{eq: multiclass} can be found in the SM~\ref{Appendix proof}.

In practice, we obtain feature sets $\{e^{j,t}_{i}\}^{n_t}_{i=1}$ with all available encoders $\{E_j(\cdot)\}^{t-1}_{j=1}$ for task collection ${\cal T}$, and calculate $S(j,t)$ for a new task $D_t$. Then we select $\mathcal{T}_a$ with $a = {\rm argmin}_j \ \{ S(j,t) \}^{t-1}_{j=1}$.
Further, for each data point $i$ in $D^t$, we obtain $p(u_i = j|\mathbf{x}_i)$, for $j=1,\ldots,t-1$ using the ELBO to evaluate $p({\bf x}_i|\phi_j)$.
Then we can estimate the probability of consistency between $\mathcal{T}_t$ and $\mathcal{T}_j$ via \eqref{eq: j(u_n = k|x_n)} and select $\mathcal{T}_b$ with $b = {\rm argmax}_j \ \{p(\mathcal{P}_t = \mathcal{P}_j)\}^{t-1}_{j=1}$.
Finally, if $\mathcal{T}_a = \mathcal{T}_b$ then $\mathcal{T}_a$ is a similar task, otherwise, $\mathcal{T}_t$ is a dissimilar task. 
The SDR framework is illustrated in Figure~\ref{fig:SDR Diagram}, from which we see that for the collection of currently learned tasks, we keep a \emph{knowledge repository} containing (for the tasks seen so far ${\cal T}$) $E_0(\cdot)$, $\{E_j(\cdot)\}_{j=1}^{t-1}$, $\{G_j(\cdot)\}_{j=1}^{t-1}$, $\{f_t(\cdot)\}_{j=1}^{t-1}$, \emph{i.e.}, the backbone encoder, task-specific encoder-decoder pairs, and the classification heads, respectively.
For a more detailed view of the SDR framework see Algorithm \ref{alg:SDR} in the SM.

\section{Experiments}
\label{Experiments}
We test our method on several image benchmark datasets, including \texttt{EMNIST}, \texttt{CIFAR10}, \texttt{CIFAR100}. Next we describe how to construct dissimilar and similar tasks.

For \texttt{CIFAR100}, we split all the classes into 20 tasks $\{T_{k}\}^{20}_{k=1}$ with 5 classes for each task; then we divide data for each class evenly into two splits for every task to create the final dataset $\{T_{k, r}\}^{2}_{r=1}, k \in [1, 20]$. For instance, $T_{1, 1}$ and $T_{1, 2}$ have the exact same classes, but the images come from different splits of the same classes. In general, $T_{k,r}, \forall r$ are similar tasks for the same $k$.

We can create CL task sequences for \texttt{CIFAR10} and \texttt{EMNIST} in the same way as for \texttt{CIFAR100}. \texttt{CIFAR10} has 10 classes in total, so we can evenly split the entire dataset into 5 tasks $\{T_{k}\}^{5}_{k=1}$ with 2 classes for each task, and then create the mixed sequence dataset $\{T_{k, r}\}^{2}_{r=1}, k \in [1, 5]$. For \texttt{EMNIST} (balanced version), to create a 100-task sequence, we divide 47 classes into 10 tasks $\{T_{k, r}\}^{10}_{r=1}, k \in [1, 10]$ with 5 classes for  $\{T_{k, r}\}^{10}_{r=1}, k \in [1, 9]$, and 2 classes for $\{T_{k, r}\}^{10}_{r=1}, k = 10$. Additional details of the datasets can be found in Tables~\ref{table:datasets} and \ref{table: task creation} in the SM \ref{Appendix dataset details}.

\vspace{-5pt}
\subsection{Model Architecture and Experimental Details}
\vspace{-3pt}
We start the CL experiments with a \emph{backbone} $E_0(\cdot)$ and the models for three dissimilar tasks trained in the repository. The system seeks to identify if a task is new or seen to the CL system for $N$ new tasks, with $N+3$ being the total number of tasks. Different task ordering in the sequence may lead to different results for the system. To show the robustness of our method to the effects of task ordering, all the experiments are repeated with five randomly permuted task sequences.

We use a pretrained ResNet18 \cite{He2016DeepRL} model on ImageNet \cite{deng2009imagenet} as our global backbone model $E_0(\cdot)$, and for each new task $\mathcal{T}_t$, we train a set of task-specific parameters for $E_t(\cdot)$, $G_t(\cdot)$ and $f_t(\cdot)$.
The autoencoding pair for the distributional consistency estimator is implemented with a standard variational autoencoder (VAE) \cite{Bank2020Autoencoders}. To further reduce memory footprint, we also apply the feature transformation approach discussed in Section~\ref{task-specific-encoder} to the convolutional layers in VAE models for the experiments with \texttt{CIFAR10} and \texttt{CIFAR100}.
We start with a VAE model pretrained on ImageNette \cite{imagenette}, a subset of ImageNet, and for each new task $\mathcal{T}_t$, similar to $E_t(\cdot)$, we train a set of task-specific parameters for each $G_t(\cdot)$.
More architecture and experimental details can be found in the SM \ref{Appendix experimental details}.

We evaluate our method using five performance characteristics.
Due to the nature of supervised learning tasks, we measure the overall \textbf{average accuracy} (Acc) after learning all the tasks in the sequence. For each task, we report the test-set accuracy. The highlight of our work is the similar task detection module. Therefore, it is crucial to show how well the framework can identify correctly whether a task is new or similar to a previously learned task. We analyze the ratio of the number of \textbf{correct identification} (short for correct) to $N$ new tasks. As for incorrect predictions (identification mistakes), we also evaluate the percentage of times that the system \textbf{misses} (does not recognize) the similar task previously seen in the past, and that the \textbf{incorrect identification} (short for incorrect) to reuse to $N$ new tasks.\
Further, we report the \textbf{overall memory usage} (in MB) by the end of training.

\vspace{-5pt}
\subsection{Baselines}
\vspace{-3pt}
The CL baseline methods to which we compare are: 
($a$) \textbf{Single Model per task:} Training a model separately for each task without finetuning, which means it will not suffer from CF.
($b$) \textbf{Optimal:} The optimal situation for our method, where the system always selects the right model to reuse for tasks in the sequence or trains a new set of VAE parameters and classifier for the new task. This serves as a strong upper bound for the proposed approach.
($c$) \textbf{Finetune:} Sequentially training a single model on all the tasks without handling forgetting issues.
($d$) \textbf{New-Head:} A single model with a shared feature extractor for all the tasks and a new classification head for each individual task.
($e$) \textbf{Online EWC} \cite{Schwarz2018ProgressC}: Training with a regularization term added to the loss function for a single model, so that the changes to important parameters are penalized during training for later tasks.

($f$) \textbf{Experience Replay} \cite{Rolnick2019ExperienceRF}: Finetuning with the subset of exemplars (10 for each class) saved for each task for rehearsal purpose.
($g$) \textbf{Deep Generative Replay (DGR)} \cite{Shin2017ContinualLW}: Training a generative model to replay the images. Following \cite{Shin2017ContinualLW}, the replayed images were labeled with the most likely category predicted by a copy of the main model stored after training on the previous task, {\em i.e.}, hard targets.
($h$) \textbf{HAT} \cite{Serr2018OvercomingCF}: Learning an attention mask over the parameters in backbone to prevent forgetting. For fair comparison, we implement HAT with a wider version of the original AlexNet model used in \cite{Serr2018OvercomingCF}.
($i$) \textbf{CAT} \cite{mixedsequence}: This method considers a similar CL setting as ours. Additional baseline architecture details can be found in the SM \ref{Appendix experimental details}.

\setlength{\tabcolsep}{4pt}
\begin{table}[t]
\begin{center}
\caption{Task similarity prediction performance during continual learning. Results are averages over five randomly permuted CL sequences.}
\label{table:task similarity prediction}
\vspace{1mm}
\begin{tabular}{lccc}
\hline\noalign{\smallskip}
&\multirow{2}{*}{Correct} & \multicolumn{2}{c}{Mistake}\\
\noalign{\smallskip}
SDR (EFT for $E\&G$)& & Miss & Incorrect \\
\noalign{\smallskip}
\hline
\noalign{\smallskip}
\texttt{CIFAR10} & 88.6$\%$ & 2.9$\%$ & 8.6$\%$\\
\texttt{CIFAR100} & 89.7$\%$ & 4.3$\%$ & 5.9$\%$\\
\texttt{EMNIST} & 93.8$\%$ & 0.2$\%$ & 6.0$\%$\\
\hline
\noalign{\smallskip}
SDR (EFT for $E\&G$) \\
\noalign{\smallskip}
\hline
\noalign{\smallskip}
\texttt{CIFAR10} & 97.1$\%$ & 2.9$\%$ & 0.0$\%$\\
\texttt{CIFAR100} & 88.1$\%$ & 3.8$\%$ & 8.1$\%$\\
\hline
\end{tabular}
\end{center}
\end{table}
\vspace{-1em}

\subsection{Results}
\vspace{-3pt}
We first show the performance of task similarity prediction module in Table \ref{table:task similarity prediction}. As a key component in our system, it demonstrates high prediction accuracy across all benchmark datasets. The performance remains competitive for \texttt{CIFAR10} and \texttt{CIFAR100} even after we further reduce the memory usage by applying EFT feature transformation to the generators $G(\cdot)$. We also note that, according to our experimental results, the ImageNette-pretrained VAE is not suitable as a backbone for EMNIST data, thus we do not report the corresponding results.

\begin{figure}[t]
\centering

\includegraphics[width=\textwidth]{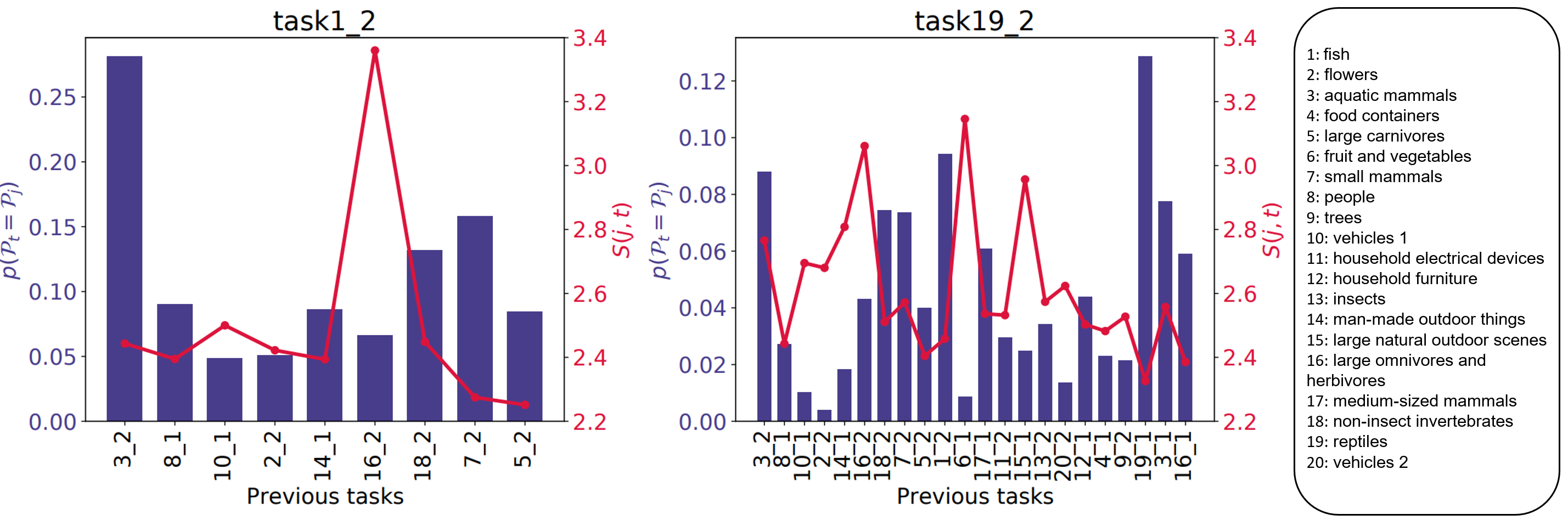}
\vspace{-.5em}
\caption{Examples from the same permuted CIFAR100 task sequence. The left y-axis shows  $p(\mathcal{P}_t = \mathcal{P}_j)$ of the current task with previous tasks. The right y-axis shows the value of $S(j, t)$ for each previous task $j$ with current task $t$. The x-axis represents all the previously identified dissimilar tasks by the system. On the right, we show a list of the 20 superclass labels, the indexes of which correspond to $k$ in task $T_{k,r}$.
}
\label{fig:example}
\end{figure}

In Figure \ref{fig:example}, we also present two task similarity prediction examples from the same permuted sequence of \texttt{CIFAR100}. The left figure shows the case where the system successfully identifies $\mathcal{T}_t$ as a new task. Moreover, the figure on the right shows the case where the system successfully identifies $\mathcal{T}_t$ as a similar task from the past. Task $T_{1,2}$ images are the most similar to task $T_{3,2}$ based on $p(\mathcal{P}_t = \mathcal{P}_j)$.
Task $T_{1,2}$ consists of images in superclass ``fish'' from \texttt{CIFAR100} while task $T_{3,2}$ includes images from superclass ``aquatic mammals''. Both tasks have a substantial amount of images with a blue ocean background, and dolphins and whales look like sharks. Regardless, our similar task identification module realizes that features extracted with task $T_{3,2}$ are not the most consistent with the labels given the current learning knowledge. Concerning task $T_{19,2}$, from all 22 tasks kept in the knowledge repository, the system recognizes that the images for the current task are the most similar to $T_{19,1}$.
Hence, the features extracted with the model for task $T_{19,1}$ can describe the labels the best.

We also compare SDR with several CL baselines in Table~\ref{table:compare with baselines}, in terms of continual learning performance and memory usage. Though the naive Single Model per task achieves highest average accuracy for all datasets, its memory consumption is comparatively too high. While maintaining a comparable accuracy performance to Single Model per task and Optimal, we successfully reduce the memory footprint. This advantage is even more noticeable when the number of tasks increases and more similar tasks are in the sequence, as revealed in the EMNIST experiment. Notably, since finetuning with wrongly identified tasks on previous encoder $E(\cdot)$ may lead to overall deterioration of predictive performance, we freeze the parameters for tasks in the repository, which yields no negative backward transfer. In practice, if desired, we can trade-off memory and accuracy by initializing a new set of parameters weights from a previous similar task, which will lead to positive forward transfer. However, this is left as interesting future work.

\setlength{\tabcolsep}{4pt}
\begin{table}[t]
\begin{center}
\caption{Comparing SDR with continual learning baselines. Results are averages over five randomly permuted CL sequences.}
\label{table:compare with baselines}
\small
\begin{tabular}{lccccccccc}
\hline\noalign{\smallskip}
Method & \texttt{CIFAR10} & (10 tasks) & & \texttt{CIFAR100} &(40 tasks) & & \texttt{EMNIST} &(100 tasks)\\
\noalign{\smallskip}
 & Acc($\%$) & Mem(MB) & & Acc($\%$) & Mem(MB) & & Acc($\%$) & Mem(MB)\\
\noalign{\smallskip}
\hline
\noalign{\smallskip}
Optimal (EFT for $E$)  & 88.45 & 76.2 & & 68.30 & 169.2 & & 97.41 & 107.0\\
Optimal (EFT for $E$ and $G$) & 88.45 & 64.7 & & 68.30 & 114.0 & & - & -\\
Single Model per task & 92.28 & 453.0 & &72.97 & 1812.0 & & 97.39 & 4530.0\\
[5pt]
Finetune & 50.92 & 45.3 & & 21.94 & 45.3 & & 32.84
& 45.3\\
New-Head & 56.10 & 45.3 & & 23.22 & 45.7 & & 36.93 & 46.3\\
ER & 78.22 & 55.7 & &24.83 & 61.2 & & 43.36 & 71.0\\
DGR & 63.70 & 95.0 & & 21.79 & 95.0 & & 25.16 & 95.0\\
Online EWC & 79.54 & 140.4 & &43.31 & 140.4 & & 89.89 & 140.4\\
CAT & 79.52 & 210.8 & &50.20 & 220.8 & & 93.27 & 200.0 \\
HAT & 86.63 & 104.7 & & 37.09 & 108.7 & & 93.76 & 114.7\\
[5pt]
SDR (EFT for $E$) & 87.25 & 77.4 & &\textbf{67.17} & 176.8 & & \textbf{96.26} & 109.0\\
SDR (EFT for $E$ and $G$) & \textbf{88.64} & 65.4 & &66.94 & 119.2 & & - & -\\
\hline
\end{tabular}
\end{center}
\end{table}
\setlength{\tabcolsep}{1.4pt}
\vspace{-5pt}
\section{Conclusion and Future Work}
\label{Conclusion}
In this study, we explored a practical TCL context where tasks are not always distinct from each other. We propose a mechanism that allows for the identification of previously seen tasks to avoid repetitive training; as a means to address memory issues with parameter expansion methods for TCL. The proposed mechanism is analyzed across several image benchmarks in terms of identifying reusable modules (encoders and generators) from previous experience. Our experimental results showed promising performance even without any regularization or special modulation to the predictive models. For future work, we intend to explore the possibility of integrating our task similarity detection module in even more realistic Cl settings, where there is class overlap between tasks.

\clearpage
\bibliographystyle{unsrt} 
\bibliography{main_paper}

\medskip
\clearpage
\section*{Checklist}
%

\begin{enumerate}
\item For all authors...
\begin{enumerate}
  \item Do the main claims made in the abstract and introduction accurately reflect the paper's contributions and scope?
    \answerYes{}
  \item Did you describe the limitations of your work?
    \answerYes{} The continual learning scenario we discuss in the paper is just one practical case of real-world settings. In Section \ref{Conclusion}, we claimed that we are going to explore even more practical settings where there is class overlap between tasks.
  \item Did you discuss any potential negative societal impacts of your work?
    \answerNo{}
  \item Have you read the ethics review guidelines and ensured that your paper conforms to them?
    \answerYes{}
\end{enumerate}

\item If you are including theoretical results...
\begin{enumerate}
  \item Did you state the full set of assumptions of all theoretical results?
    \answerYes{} The assumptions of theoretical results can be found in Section \ref{background}.
        \item Did you include complete proofs of all theoretical results?
    \answerYes{} It is shown in the SM~\ref{Appendix proof}.
\end{enumerate}

\item If you ran experiments...
\begin{enumerate}
  \item Did you include the code, data, and instructions needed to reproduce the main experimental results (either in the supplemental material or as a URL)?
    \answerNo{} The code will be released upon acceptance.
  \item Did you specify all the training details (e.g., data splits, hyperparameters, how they were chosen)?
    \answerYes{} Model architecture and experimental details, including data description, hyperparameter selection are provided in the SM \ref{Appendix dataset details} and \ref{Appendix experimental details}.
        \item Did you report error bars (e.g., with respect to the random seed after running experiments multiple times)?
    \answerYes{} The corresponding performance is provided in the SM.
        \item Did you include the total amount of compute and the type of resources used (e.g., type of GPUs, internal cluster, or cloud provider)?
     \answerNo{} We have not, however, these details will be included in the final version.
\end{enumerate}

\item If you are using existing assets (e.g., code, data, models) or curating/releasing new assets...
\begin{enumerate}
  \item If your work uses existing assets, did you cite the creators?
    \answerYes{}
  \item Did you mention the license of the assets?
    \answerNo{} The code reference is not mentioned in the main paper, but we will provide the license of the assets in our code base acknowledgement section.
  \item Did you include any new assets either in the supplemental material or as a URL?
    \answerNo{} The CL tasks sequences will be released with the source code.
  \item Did you discuss whether and how consent was obtained from people whose data you're using/curating?
    \answerNo{} The data we are using in the paper is open-source.
  \item Did you discuss whether the data you are using/curating contains personally identifiable information or offensive content?
    \answerNo{} There is no personally identifiable information or offensive content in the data we are using.
\end{enumerate}

\item If you used crowdsourcing or conducted research with human subjects...
\begin{enumerate}
  \item Did you include the full text of instructions given to participants and screenshots, if applicable?
    \answerNA{}
  \item Did you describe any potential participant risks, with links to Institutional Review Board (IRB) approvals, if applicable?
    \answerNA{}
  \item Did you include the estimated hourly wage paid to participants and the total amount spent on participant compensation?
    \answerNA{}
\end{enumerate}

\end{enumerate}

\clearpage

\appendix
\input{Supplemental_Materials}

\end{document}

%% file: Supplemental_Materials.tex
\section{Appendix}
\label{Appendix}

\subsection{Experimental Details}
\label{Appendix experimental details}
In all the experiments except for methods CAT and HAT, we use a pretrained ResNet18 as the backbone of the classification models. To adapt the same pretrained backbone model for EMNIST, which contains images of one channel, we first transform the grey-scale images into three-channel images. For a fair comparison, we adopt the efficient feature transformation technique in Optimal, Finetune, New-Head, and SDR. Serving as a benchmark, Single Model per task is trained with a full ResNet18 for each task. They all use the pretrained ResNet18 on \texttt{ImageNet} as a backbone. For CAT, we show the results with a 2 fully connected layer network used in the original paper.
Following the setting in CAT paper \cite{mixedsequence}, the embeddings for the hidden and final layer in the knowledge base have a dimension of 2000; the task ID embeddings also have 2000 dimensions. For HAT, we switched to a wide version of AlexNet model as backbone. Specifically, each layer in the wide version has twice the number of nodes compared to each layer in the original version \cite{Serr2018OvercomingCF}. For DGR, we follow the setting in \cite{vandeVen2018GenerativeRW} for the generative model, which is a symmetric VAE with 3 fully connected layers (fc1000 - fc1000 - fc100). As for the VAE models in SDR, an architecture of 4 convolutional layers and 1 linear layer is used for both encoder and decoder.

All the classification models (when feature extractors are newly trained) are trained with a batch size of 128 for 100 epochs, with a starting learning rate of 0.01 reduced with a factor of 0.1 at epoch 50. The only data augmentation/transformation used for the datasets (except for EMNIST, which is converted from one-channel to three-channels) is normalization. 
When the feature extractor is reused, the classification head only needs to be trained for 2 epochs with a learning rate of 0.001. As for VAE models trained in SDR, they are trained with a batch size of 64 and a learning rate of 0.0001, for 2000 epochs with early stopping criteria.

Here we provide a detailed view of the SDR framework in \ref{alg:SDR}.

\begin{algorithm}
\caption{SDR Framework}\label{alg:SDR}
\KwData{$D_{t} = \{\mathbf{x}_{i}^{t}, y^{t}_{i}\}^{n_t}_{i=1}, t\in[1, N]$; (Only $D_t$ available at time $t$)}
\KwResult{$\{E_{m}\}^{C}_{m=1}$, $\{G_{m}\}^{C}_{m=1}$, with $C$ being the number of unique tasks recognized by the system}
\BlankLine
\emph{Starting with:}$E_0$, $\{E_{m}\} \gets \{E_{m}\}^{3}_{m=1}$;\; $G_0$ (might or might not be available), $\{G_{m}\} \gets \{G_{m}\}^{3}_{m=1}$\; 
\BlankLine
\For{$t\leftarrow 1$ \KwTo $N$}{
    \For{$j \leq t-1$}{
    \For{$i \leftarrow 1$ \KwTo $n^t$}{
        $p(u_i=j|\mathbf{x}_i)$  \tcc*[r]{Distributional Consistency Estimator}
    }
    $p(\mathcal{P}_t = \mathcal{P}_j) = \mathbb{E}_{\mathbf{x}_i \in \mathcal{D}_t}[p(u_i = j|\mathbf{x}_i)]$ 
    }
    $\mathcal{T}_a = \arg \max_j \{p(\mathcal{P}_t = \mathcal{P}_j)\}^{t-1}_{=1}$ \; 
    
    \For{$j \leq t-1$}{
    $\mathbf{H}^{j, t}_{ik} = \frac{{\mathbf{e}^{j, t}_{i}}^\top \mathbf{e}^{k, t}_{k}(\pi - \arccos({\mathbf{e}^{j, t}_i}^\top \mathbf{e}^{j, t}_{k}))}{2\pi}$,\; 
    $\mathbf{A}^{j, t} = \mathbf{y}^\mathsf{T} (\mathbf{H}^{j, t})^{-1}\mathbf{y}$
    
    $S(j, t) \gets \sqrt{\frac{2 ||{\mathbf{A}^{j, t}}^\top \mathbf{A}^{j, t}||^2_2} {n_t}}$ \;   \tcc*[r]{Predictor-label Association Analysis}
    }
    $\mathcal{T}_b = \arg \min_{j} \{S(j, t)\}^{t-1}_{j=1}$ \; 
    
    \eIf(\\reuse feature encoder and VAE model from task $\mathcal{T}_a$ for task $\mathcal{T}_t$, \\ only train a new classification head $f_t$ for $\mathcal{T}_t$){$\mathcal{T}_a == \mathcal{T}_b$}{
        $E_{t} \gets E_{a}$\; 
        
        $G_{t} \gets G_{a}$\; 
    }
    (train new models for task $\mathcal{T}_t$){
        \Repeat{$\theta_{E_{t}}$ converges}{
        $\mathbf{B} \gets$ random sampled mini-batch from $D_t$
        
         $\mathbf{g} = \nabla_{\theta_{E_{t}}} \mathcal{L}(\phi_{E_{t}}, \mathbf{B})$
         
         $\theta_{E_{t}} \gets AdamOptimizer(\theta_{E_{t}}, \mathbf{g})$\; 
        }
        \Repeat{$\psi_{G^t}, \phi_{G^t}$ converges}{
            $\mathbf{B} \gets$ random sampled mini-batch from $D_t$
            
            $\mathbf{g} = \nabla_{\psi_{G_t}, \phi_{G_{t}}} \mathcal{L}(\psi_{G_t}, \phi_{G_{t}}, \mathbf{B})$
            
            $\psi_{G_t}, \phi_{G_{t}} \gets AdamOptimizer(\psi_{G_t}, \phi_{G_{t}}, \mathbf{g})$\; 
        }
    }
}
\end{algorithm}

\subsection{Dataset Details}
In Table \ref{table:datasets}, we provide the number of tasks in the sequences created with each dataset, the number of classes for each task, and the size of each data split.
\label{Appendix dataset details}
\setlength{\tabcolsep}{4pt}
\begin{table}[ht]
\begin{center}
\caption{Details of the Datasets}
\label{table:datasets}
\begin{tabular}{ccccccc}
\hline\noalign{\smallskip}
Dataset & \# tasks & \# classes/tsk & training/tsk & validation/tsk & testing/tsk\\
\noalign{\smallskip}
\hline
\noalign{\smallskip}
\texttt{CIFAR10} & 10& 2 & 4500 & 500 & 1000\\
\texttt{CIFAR100} & 40& 5 & 1125 & 125 & 250\\
\texttt{EMNIST} & 100 & 5 or 2 & 1080 or 432 & 120 or 48 & 200 or 80\\
\hline
\end
{tabular}
\end{center}
\end{table}
\setlength{\tabcolsep}{1.4pt}

\setlength{\tabcolsep}{4pt}
\begin{table}
\begin{center}
\caption{Details for Classes in Each Task}
\label{table: task creation}
\begin{tabular}{lll}
\hline\noalign{\smallskip}
Dataset & $k$ & Classes \\
\noalign{\smallskip}
\hline
\noalign{\smallskip}
\texttt{CIFAR10} & 1& ['airplane', 'bird'] \\
        & 2& ['automobile', 'truck'] \\
        & 3& ['cat', 'dog'] \\
        & 4& ['deer', 'horse'] \\
        & 5& ['frog', 'ship'] \\
\texttt{CIFAR100} & 1& ['aquarium fish', 'flatfish', 'ray', 'shark', 'trout'] \\
        & 2& ['orchid', 'poppy', 'rose', 'sunflower', 'tulip'] \\
        & 3& ['beaver', 'dolphin', 'otter', 'seal', 'whale'] \\
        & 4& ['bottle', 'bowl', 'can', 'cup', 'plate'] \\
        & 5& ['bear', 'leopard', 'lion', 'tiger', 'wolf'] \\
        & 6& ['apple', 'mushroom', 'orange', 'pear', 'sweet pepper']  \\
        & 7& ['hamster', 'mouse', 'rabbit', 'shrew', 'squirrel'] \\
        & 8& ['baby', 'boy', 'girl', 'man', 'woman'] \\
        & 9& ['maple tree', 'oak tree', 'palm tree', 'pine tree', 'willow tree'] \\
        & 10& ['bicycle', 'bus', 'motorcycle', 'pickup truck', 'train'] \\
        & 11& ['clock', 'keyboard', 'lamp', 'telephone', 'television'] \\
        & 12& ['bed', 'chair', 'couch', 'table', 'wardrobe'] \\
        & 13& ['bee', 'beetle', 'butterfly', 'caterpillar', 'cockroach']\\
        & 14& ['bridge', 'castle', 'house', 'road', 'skyscraper'] \\
        & 15& ['cloud', 'forest', 'mountain', 'plain', 'sea'] \\
        & 16& ['camel', 'cattle', 'chimpanzee', 'elephant', 'kangaroo']  \\
        & 17& ['fox', 'porcupine', 'possum', 'raccoon', 'skunk'] \\
        & 18& ['crab', 'lobster', 'snail', 'spider', 'worm'] \\
        & 19& ['crocodile', 'dinosaur', 'lizard', 'snake', 'turtle'] \\
        & 20& ['lawn mower', 'rocket', 'streetcar', 'tank', 'tractor'] \\
\texttt{EMNIST}  & 1& ['0', '1', '2', '3', '4'] \\
        & 2& ['5', '6', '7', '8', '9'] \\
        & 3& ['A', 'B', 'D', 'E', 'F'] \\
        & 4& ['G', 'H', 'N', 'Q', 'R'] \\
        & 5& ['T', 'a', 'b', 'c', 'd'] \\
        & 6& ['e', 'f', 'g', 'h', 'i'] \\
        & 7& ['j', 'k', 'l', 'm', 'n'] \\
        & 8& ['o', 'p', 'q', 'r', 's'] \\
        & 9& ['t', 'u', 'v', 'w', 'x'] \\
        & 10& ['y', 'z'] \\
\hline
\end
{tabular}
\end{center}
\end{table}
\setlength{\tabcolsep}{1.4pt}

\subsection{Proof}
\label{Appendix proof}
In Section \ref{sec:A.3.1} to \ref{sec:A.3.3}, we provide theoretical proofs for our similarity metric, while in Section \ref{sec:A.3.4} we show empirical proofs.
\subsubsection{Setting} \label{sec:A.3.1}
Given a two-layer ReLU activated neural network with $m$ neurons in the hidden layer and $c$ neurons in the output layer:
\begin{align}
     f_{\mat{W}, \mat{A}}(\vect{x}) = \frac{1}{\sqrt{m}}\sum^{m}_{r=1} \vect{a}_r \sigma(\vect{w}_{r}^{\top}\vect{x}),
\end{align}
where $\vect{x} \in \R^d$ is an input,  $\mat{W} = (\vect{w}_1, \cdots, \vect{w}_m) \in \R^{d \times m}$ are the weight vectors in the first layer and $\mat{A} = (\vect{a}_1, \cdots, \vect{a}_m)^{\top} \in \R^{c \times m}$ are weight vectors in the second layer, with $c$ being the number of classes.

We train the neural network by \emph{randomly initialized gradient descent} (GD) on the quadratic loss over data. In particular, we first initialize the parameters randomly:\footnote{For each $r\in [1,m]$, randomly pick sum $s_r$ of the $r$-th row $\vect{a}_r$ from  $\text{unif}(\{-1,1\})$. For the first $c-1$ neurons, we randomly sample $a_{z,r} \sim \text{unif}([-1, 1]), \forall z = 1,\cdots, c-1$, the last neuron is set to satisfy the column sum, {\it i.e.}, $a_{c, r} = s_r - \sum^{c-1}_{z=1} a_{z, r}$.}
\begin{align}\label{eq:random initialization}
    \vect{w}_r (0) \sim \cN (\vect{0}, \kappa^2 \mat{I}), \qquad \sum^{c}_{z=1} a_{z, r} \sim \text{unif}(\{-1, 1\}), \qquad \forall r \in [m],
\end{align}
where $0 < \kappa \leq 1$ controls the magnitude of initialization.

We want to minimize quadratic loss of
\begin{align}
    \cL(\vect{W}) = \frac{1}{2N} \sum^{N}_{i=1} \sum^{c}_{z=1} (y_{iz} - f_{\mat{W}, \mathbf{a}_z}(\vect{x}_i))^2,
\end{align}
through GD by fixing the second layer $\mathbf{A}$ and only optimizing the first layer $\mathbf{W}$
\begin{align} \label{eq: gradient of L}
    \frac{\partial \cL(\mat{W})}{\partial \vect{w}_r} = \frac{1}{2N}
    \sum^{c}_{z=1} a_{z, r} \sum^{N}_{i=1} (f_{\mat{W}, \vect{a}_z}(\vect{x}_i) - y_{iz}) \mathbb{I} \{\vect{w}_r^{\top} \vect{x}_i \geq 0\} \vect{x}_i, \qquad \forall r \in [m].
\end{align}

Define $\hat{y}_{iz} = f_{\mat{W}, \vect{a}_z}(\vect{x}_i)$, {\it i.e.}, the network's prediction on the $i$-th input that belongs to class $z$, and \eqref{eq: gradient of L} can be rewritten as:
\begin{align}
    \frac{\partial \cL(\mat{W})}{\partial \vect{w}_r} = \frac{1}{2N}
    \sum^{c}_{z=1} a_{z, r} \sum^{N}_{i=1} (\hat{y}_{iz} - y_{iz}) \mathbb{I}_{r,i} \vect{x}_i, \qquad \forall r \in [m],
\end{align}
where $\mathbb{I}_{r, i} = \mathbb{I}\{\mathbf{\vect{w}^{\top}_{r} \vect{x}_i} \geq 0\}$. We also define $\mathbf{Z}$ as:
\begin{align}
    \mat Z = \frac{1}{\sqrt{m}} \begin{bmatrix}
        \mathbb{I}_{1,1} (\sum^{c}_{z=1} a_{z,1}) \vect{x}_1 & \cdots & \mathbb{I}_{1,1} (\sum^{c}_{z=1} a_{z,1}) \vect{x}_n \\
        \vdots & \ddots & \vdots \\
        \mathbb{I}_{m,1} (\sum^{c}_{z=1} a_{z, m}) \vect{x}_1 & \cdots & \mathbb{I}_{m,n} (\sum^{c}_{z=1} a_{z,m}) \vect{x}_n 
\end{bmatrix}
\end{align}

With this notation we have a more compact form of the gradient \eqref{eq: gradient of L}:
\begin{align}
    \vectorize{\vect{\nabla (\cL(\mat {W}))}} = \mat {Z}(\hat{\vect{y}} - \vect{y}).
\end{align}
$\mat H = \mat Z^\top \mat Z$,\footnote{$\mat H$ corresponds to $\mat H^\infty$ in \cite{arora2019finegrained}, which denotes the Gram matrix when GD converges, here we abuses the notation to be consist with the main paper} since we initialize the second layer with $\sum^{c}_{z=1} a_{z,r} \sim \text{unif}(\{-1, 1\})$
\begin{align}
    \mat H_{ij} &= \frac{\vect{x}_i \vect{x}_j}{m} \sum^{m}_{r=1} \mathbb{I}_{r,1} \mathbb{I}_{r,2} (\sum^{c}_{z=1} a_{z,r})^2 \\
    &= \frac{\mat {x}_i \vect{x}_j}{m} \sum^{m}_{r=1} \mathbb{I}_{r,i} \mathbb{I}_{r,j}, \qquad \forall i, j \in [n]
\end{align}
and thus we prove that we can obtain the Gram matrix $\mat H$ the same form for multi-classification setting as in binary-classification settings, and complete the proof of \eqref{eq: Our Gram Matrix}.

\subsubsection{Rademacher Complexity and Generalization}
Generalization error measures how accurate an algorithm is for predicting outcome values for previously unseen data. Specifically, given a function $f: \R^d \rightarrow \R$ and a loss function $\mathit{l}: \R \times \R\rightarrow \R$,  generalization error can be defined as the gap between the \emph{population loss} over data distribution $\cP$ and the \emph{empirical loss} over $n$ samples $\cD=\{ (\vect{x}_i, y_i)\}^{n}_{i=1}$ from $\mathcal{P}$:
\begin{align}
    \mathcal{L}_{\cP}(f) - \mathcal{L}_{\mathcal{D}}(f) = \mathbb{E}_{(\vect{x}, y)\sim \cP} [\mathit{l}(f(\vect{x}, y)] - \frac{1}{n} \sum^{n}_{i=1} \ell(f(\vect{x}_i), y_i)
\end{align}
\begin{definition}[\citep{arora2019finegrained}]
    Given $n$ samples $\cD$, the empirical Rademacher complexity of a function class $\cF$ (mapping from $\R^d$) is defined as:
    \begin{align*}
        \calR_{\cD}(\cF) = \frac{1}{n} \mathbb{E}_{\boldsymbol{\epsilon}} \bigg[\sup_{f \in \cF} \sum^{n}_{i=1} \epsilon_i f(\vect{x}_i) \bigg],
    \end{align*}
    where $\boldsymbol{\epsilon} = (\epsilon_1, \cdots, \epsilon_n)^{\top}$ contains $i.i.d$ random variables drawn from the Rademacher distribution \text{unif} (\{1, -1\}).
\end{definition}
Rademacher complexity directly gives an upper bound on generalization error:
\begin{theorem}[\citep{arora2019finegrained}]
Suppose the loss function $\mathit{l}(\cdot, \cdot)$ is bounded in $[0, c]$ and is $\rho$-Lipschitz in the first argument. Then with probability at least $1-\delta$ over sample $\mathcal{D}$ of size $n$:
\begin{align}
    \sup_{f \in \cF} \{\cL_{\cP}(f) - \cL_{\cD}(f)\} \leq 2\rho \calR_{\cD}(\cF) + 3c\sqrt{\frac{\log(2/\delta)}{2n}}
\end{align}
\end{theorem}
Therefore, as long as we can bound the Rademacher complexity of a certain function class that contains our learned predictor, we can obtain a generalization bound.

\subsubsection{Proof for Complexity Measure under Multi-class Scenario} \label{sec:A.3.3}
In this section, we show proof to the complexity measure \eqref{eq: multiclass} under multi-class classification settings. We note that most of the related proofs of complexity measure (generalization bound) under binary classification setting are provided in \cite{arora2019finegrained}. 

\begin{theorem}[\citep{Du2019GradientDP, arora2019finegrained}]
	\label{thm:ssdu-converge}
	Assume $\lambda_0 = \lambda_{\min}(\mat H^\infty) >0$. For $\delta\in(0, 1)$, if $m = \Omega\left( \frac{n^6}{\lambda_0^4 \kappa^2 \delta^3 } \right)$ and $\eta = O\left( \frac{\lambda_0}{n^2} \right)$, then with probability at least $1-\delta$ over the random initialization, we have:
	\begin{itemize}
		\item $\cL(\mat W(0)) = O(n/\delta)$;
		\item $\cL(\mat W(k+1)) \le \left( 1 - \frac{\eta \lambda_0}{2}\right)	\mathbf{\cL}(\mat W(k)),\  \forall k\ge0$.
	\end{itemize}
\end{theorem}

{\bf Analysis of the Auxiliary Sequence $\left\{\widetilde{\mat{W}}(k)\right\}_{k=0}^\infty$} \footnote{$k$ denotes the optimization step in the proof section}
Now we give a proof of  $\norm{\widetilde{\mat W}(\infty) - \widetilde{\mat W}(0)}_F = \sqrt{\norm{\vect y^\top \mat H(0)^{-1} \vect y}_F}$ as an illustration for the proof of Lemma \ref{lem:distance_bounds}.
Define $\vect v(k) = \mat Z(0)^\top \vectorize{\widetilde{\mat W}(k)} \in \R^n$.
Then from~\eqref{eqn:W_tilde_traj} we have $\vect v(0)=\vect0$ and $\vect v(k+1) = \vect v(k) - \eta \mat H(0) (\vect v(k) - \vect y)$, yielding
$\vect v(k) - \vect y = - (\mat I - \eta \mat H(0))^k \vect y$.
Plugging this back to~\eqref{eqn:W_tilde_traj} we get
$\vectorize{\widetilde{\mat{W}}(k+1)} - \vectorize{\widetilde{\mat{W}}(k)} = \eta \mat Z(0) (\mat I - \eta \mat H(0))^k \vect y$.
Then taking a sum over step $k=0, 1, \ldots$ we have
\begin{align*}
\vectorize{\widetilde{\mat{W}}(\infty)} - \vectorize{\widetilde{\mat{W}}(0)} &= \sum_{k=0}^\infty \eta \mat Z(0) (\mat I - \eta \mat H(0))^k \vect y \\
&= \mat Z(0) \mat H(0)^{-1} \vect y.
\end{align*}
The desired result thus follows:\footnote{Note that we have $\mat H(0) \approx \mat H^\infty$ from standard concentration. See Lemma C.3 in \cite{arora2019finegrained}}
\begin{align*}
\norm{\widetilde{\mat W}(\infty) - \widetilde{\mat W}(0)}_F^2
&= \vect y^\top \mat H(0)^{-1} \mat Z(0)^\top \mat Z(0) \mat H(0)^{-1} \vect y \\
&= \vect y^\top \mat H(0)^{-1} \vect y.\\
\norm{\widetilde{\mat W}(\infty) - \widetilde{\mat W}(0)}_F 
&=  \sqrt{\norm{\vect y^\top \mat H(0)^{-1} \vect y}_{F}}
\end{align*}
Lemma \ref{lem:distance_bounds}'s proof is based on the careful characterization of the trajectory of $\left\{\mat W(k) \right\}_{k=0}^\infty$ during GD.
In particular, we bound its \emph{distance to initialization} as follows:
\begin{lem}[\citep{arora2019finegrained}] \label{lem:distance_bounds} \footnote{$m$ is the number of parameters in the first layer, $\eta$ is the learning rate}
	Suppose $m \geq \kappa^{-2} \poly\left(n, \lambda_0^{-1}, \delta^{-1} \right)  $ and $\eta = O\left( \frac{\lambda_0}{n^2} \right)$.
	Then with probability at least $1-\delta$ over the random initialization, we have for all $k\geq 0$:
	\begin{itemize}
		\item $\norm{\vect w_r(k) - \vect w_r(0)}_2 =O\left( \frac{ n }{\sqrt m \lambda_0\sqrt{\delta}} \right)$ $(\forall r\in[m])$, and
		\item $\norm{\mat{W}(k)-\mat{W}(0)}_{F} \le \sqrt{\| \vect{y}^\top \left(\mat{H}^{\infty}\right)^{-1}\vect{y} \|_{F}} + O\left( \frac{  n \kappa}{\lambda_0 \delta} \right) + \frac{\poly\left(n, \lambda_0^{-1}, \delta^{-1} \right)}{m^{1/4} \kappa^{1/2}} $.
	\end{itemize}
\end{lem}
The bound on the movement of each $\vect w_r$ was proved in \cite{Du2019GradientDP}. 
The bound on $\norm{\mat{W}(k)-\mat{W}(0)}_F$ corresponds to the \emph{total movement of all neurons}.
The main idea is to couple the trajectory of $\left\{\mat W(k) \right\}_{k=0}^\infty$ with another simpler trajectory $\left\{\widetilde{\mat{W}}(k)\right\}_{k=0}^\infty$ defined as:
\begin{align}
\widetilde{\mat{W}}(0) =\,& \mat{0}, \nonumber\\
\vectorize{\widetilde{\mat{W}}(k+1)} =\,& \vectorize{\widetilde{\mat{W}}(k)} \label{eqn:W_tilde_traj} \\&- \eta\mat{Z}(0)\left( \mat{Z}(0)^\top \vectorize{\widetilde{\mat{W}}(k)}-\vect{y}\right). \nonumber
\end{align}

\begin{lem}[\citep{arora2019finegrained}]\label{lem:rad_dist_func_class} 
Given $R>0$,
with probability at least $1-\delta$ over the random initialization ($\mat W(0),\mat A)$, simultaneously for every $B>0$, the following function class
\begin{align*}
\mathcal{F}^{\mat W(0),\mat A}_{R,B}  = \{ f_{\mat W, \mat A} : \norm{\vect w_r - \vect w_r(0)}_2 \le R \, (\forall r\in[m]), \\ \norm{\mat W - \mat W(0)}_F \le B \}
\end{align*}
has empirical Rademacher complexity bounded as:
\begin{align*}
	&\calR_{\cD}\left(\cF^{\mat W(0),\mat A}_{R,B} \right)
  =  \frac{1}{n} \expect_{\boldsymbol{\epsilon} \in \{\pm1\}^n}\left[\sup_{f \in \cF^{\mat W(0),\mat A}_{R,B}}\sum_{i=1}^{n} \epsilon_i f(\vect{x}_i)\right]
	\\
	\leq \,&	\frac{B}{\sqrt{2n}} \left(1+\left( \frac{2\log \frac2\delta}{m} \right)^{1/4} \right) + \frac{2 R^2 \sqrt m}{ \kappa} + R \sqrt{2\log \frac {2}{\delta}} .
\end{align*}
\end{lem}
For the proof of Lemma \ref{lem:rad_dist_func_class}, the only difference compared to Lemma 5.4 in \cite{arora2019finegrained} is $a_r$ in \cite{arora2019finegrained} becomes $\sum^{c}_{z=1} \mat a_{r,z}$ in our setting.

Finally, combining Lemma~\ref{lem:distance_bounds} with \ref{lem:rad_dist_func_class}, we are able to conclude that the neural network found by GD belongs to a function class with Rademacher complexity at most $\sqrt{ \| \vect y^\top (\mat H^\infty)^{-1} \vect y \|_{F}/(2n)}$ (plus negligible errors).
This gives us the generalization bound in Theorem~\ref{thm:main_generalization} using Rademacher complexity.
\begin{definition}
\label{def:non-degenerate distribution}
	A distribution $\cP$ over $\R^d \times \R$ is $(\lambda_0, \delta, n)$-non-degenerate, if for $n$ i.i.d. samples $\{(\vect x_i, y_i)\}_{i=1}^n$ from $\cP$, with probability at least $1-\delta$ we have $\lambda_{\min}(\mat H^\infty) \ge \lambda_0 >0$.
\end{definition}

\begin{theorem}[\citep{arora2019finegrained}]\label{thm:main_generalization} 
	Fix 
	a failure probability $\delta \in (0, 1)$.
	Suppose our data $\mathcal{D} = \left\{(\vect x_i,  y_i)\right\}_{i=1}^n$ are i.i.d. samples from a $(\lambda_0, \delta/3, n)$-non-degenerate distribution $\cP$, and $\kappa = O\left( \frac{ \lambda_0 \delta}{ n} \right), m\ge \kappa^{-2} \poly\left(n, \lambda_0^{-1}, \delta^{-1} \right)  $.
	Consider any loss function $\ell: \R\times\R \to [0, 1]$ that is $1$-Lipschitz in the first argument such that $\ell(y, y)=0$.
	Then with probability at least $1-\delta$ over the random initialization and the training samples, the two-layer neural network $f_{\mat W(k), \mat A}$ trained by GD for $k\geq \Omega\left( \frac{1}{\eta\lambda_0} \log\frac{n}{\delta} \right)$ iterations has population loss $L_\cP(f_{\mat W(k), \mat A}) = \expect_{(\vect x, y)\sim \cP}\left[ \ell(f_{\mat W(k), \mat A}(\vect x), y) \right]$ bounded as:
	\begin{equation} \label{eqn:main_generalization}
	L_\cP(f_{\mat W(k), \mat A}) \le \sqrt{\frac{\norm{2\vect{y}^\top \left(\mat{H}^{\infty}\right)^{-1}\vect{y}}_F}{n}}  + O\left( \sqrt{\frac{\log\frac{n}{\lambda_0\delta}}{n}} \right) .
	\end{equation}
\end{theorem} 

\subsubsection{Empirical Proof for Using Complexity Measure as our Similarity Metric} \label{sec:A.3.4}

\begin{figure}[ht]
\centering
  \includegraphics[width=0.6\linewidth]{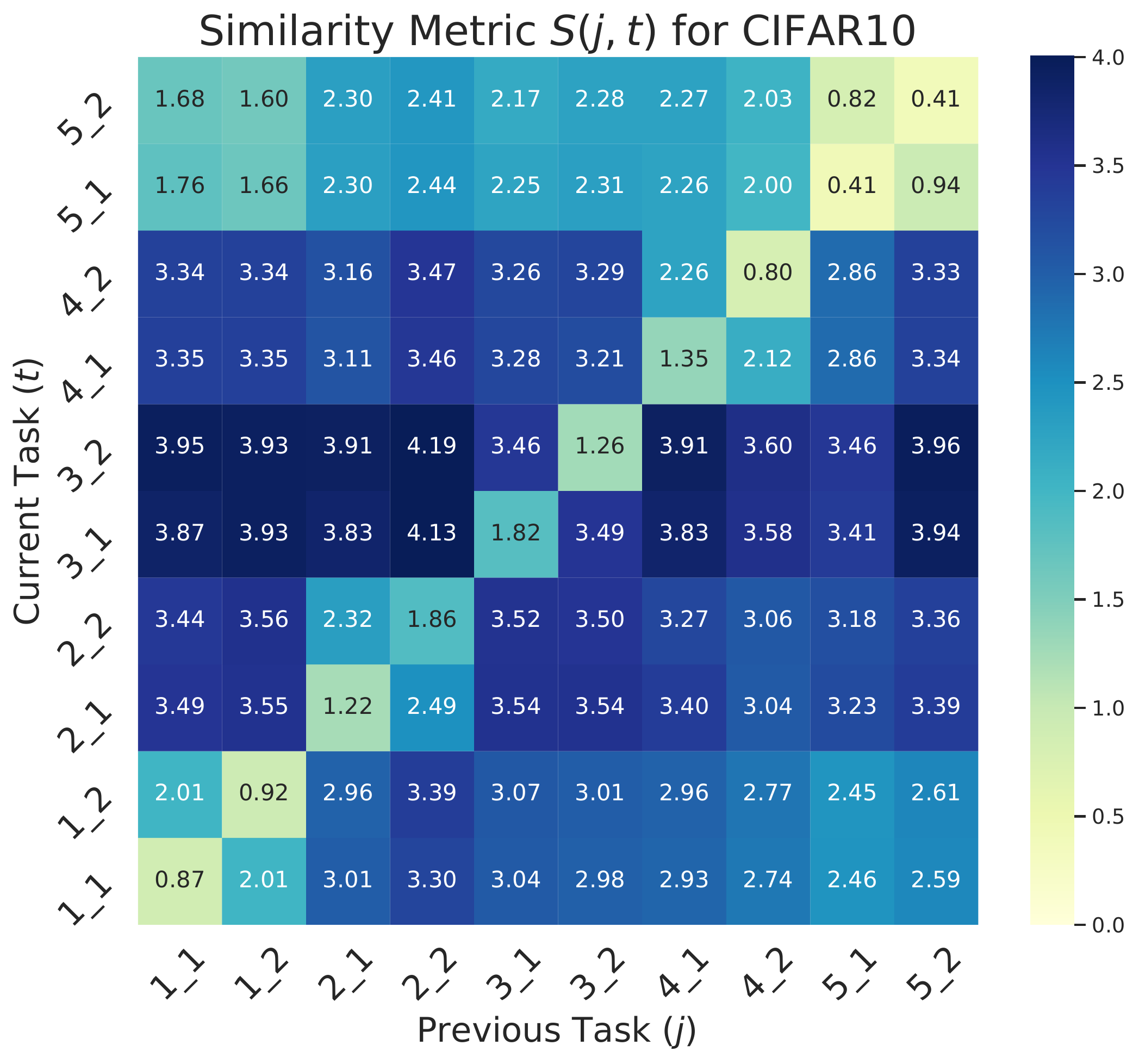}
\caption{Similarity Metric Results for \texttt{CIFAR10}}
\label{fig:similarity metric1}
\end{figure}

\begin{figure}[ht]
\centering
  \includegraphics[width=0.8\linewidth]{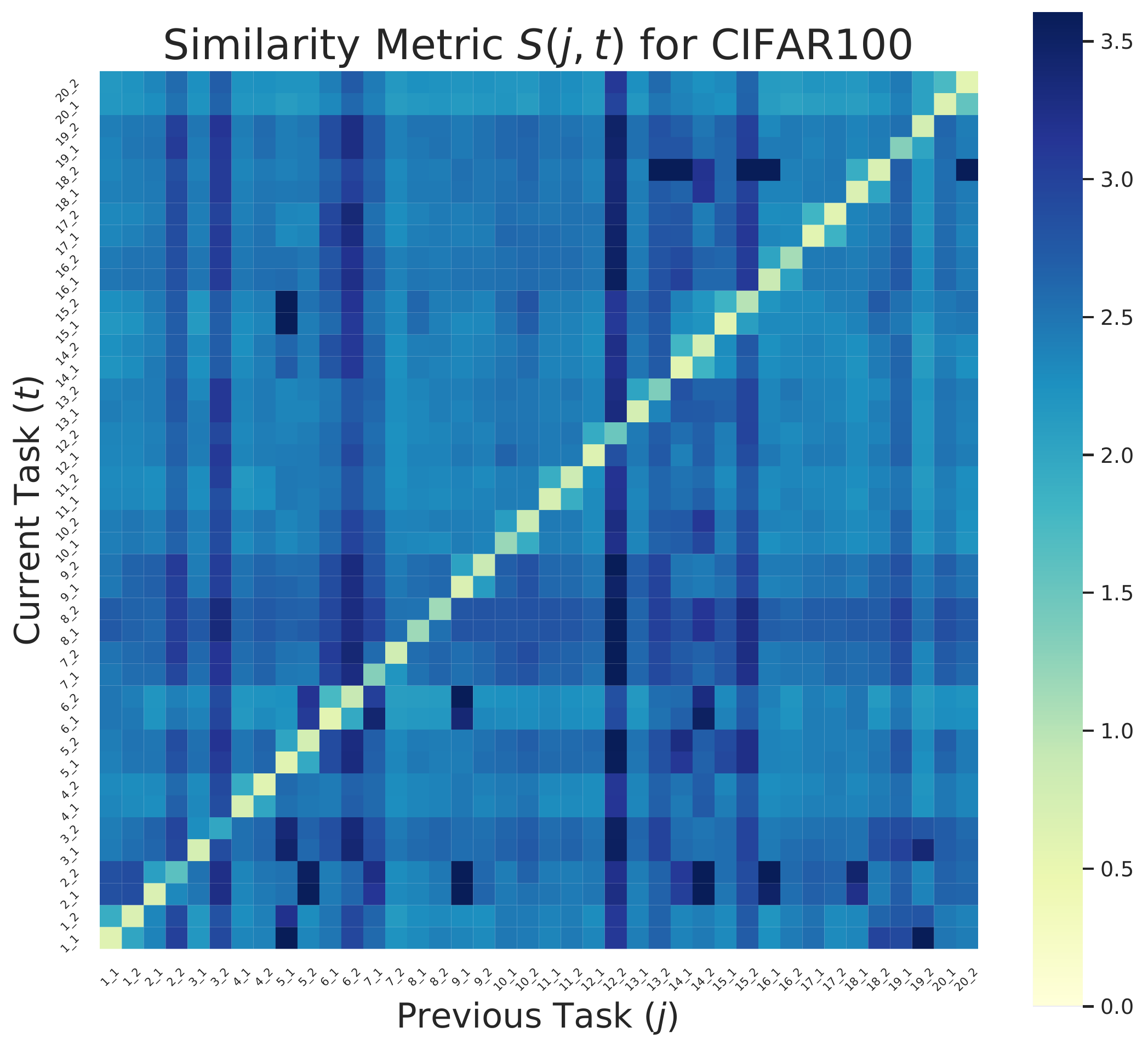}
\caption{Similarity Metric Results for \texttt{CIFAR100}}
\label{fig:similarity metric2}
\end{figure}

\begin{figure}[ht]
\centering
  \includegraphics[width=0.8\linewidth]{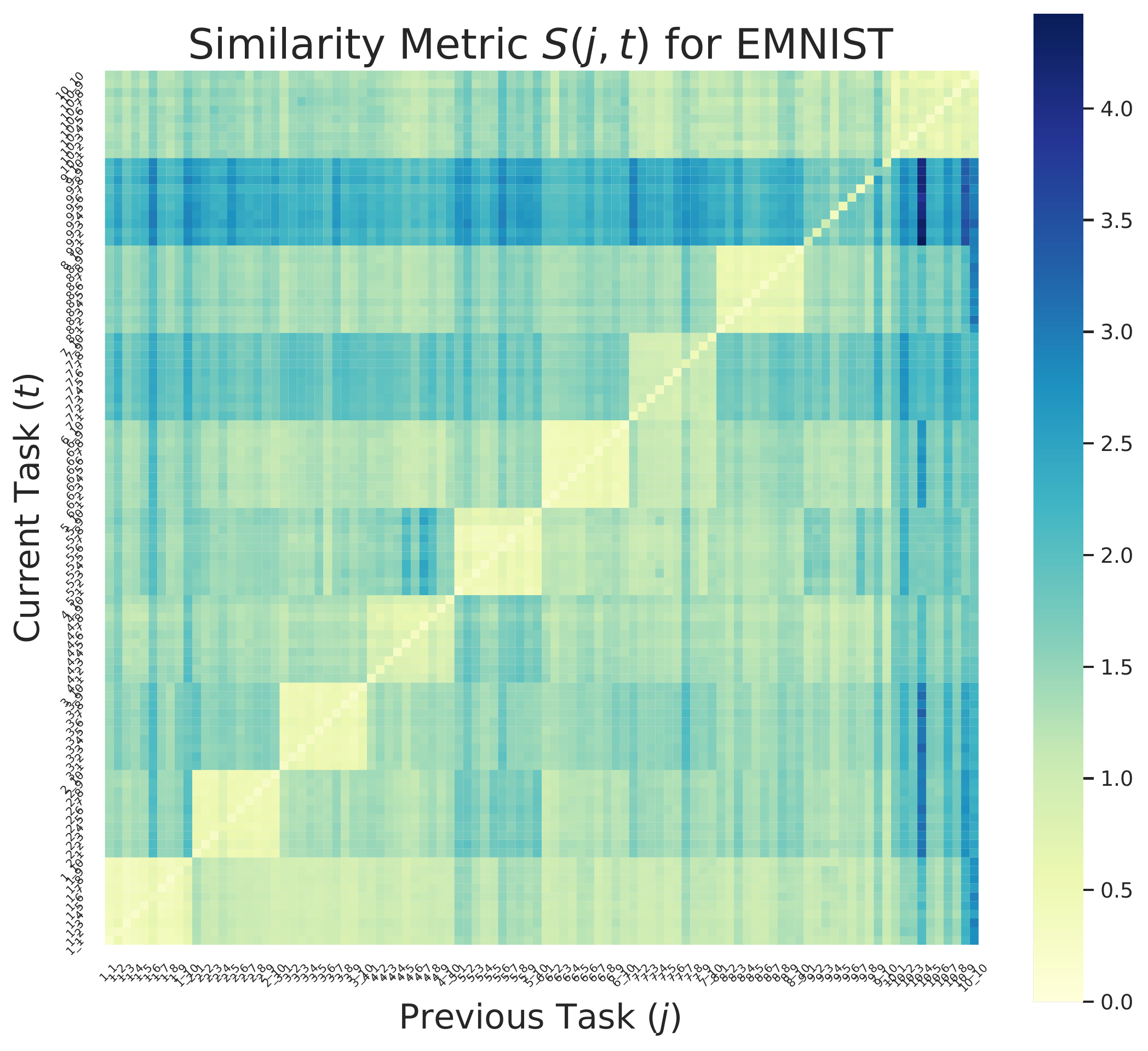}
\caption{Similarity Metric Results for \texttt{EMNIST}}
\label{fig:similarity metric3}
\end{figure}

In Figures \ref{fig:similarity metric1}, \ref{fig:similarity metric2}, \ref{fig:similarity metric3}, we intent to show that datasets encoded with feature extractors trained on similar tasks obtain relatively smaller values compared to those trained on dissimilar tasks. In the heatmaps, the y-axis represents the current tasks, while the x-axis shows the previous tasks. The value of our similarity metric is color-coded for each cell in the heatmaps. The cells with smaller values are concentrated on the right diagonals (in the light blue and green colored rectangle regions), which implies that the similarity metric values are the smallest when current tasks and previous tasks are similar. 
